\documentclass{bmvc2k}
\usepackage{marvosym}

\def\blue#1{\textcolor{blue}{#1}}

\def\blue#1{\textcolor{blue}{#1}}

\title{Mumpy: Multilateral Temporal-view Pyramid Transformer for Video Inpainting Detection}

\addauthor{Ying Zhang}{yingzhang@stu.ouc.edu.cn}{1}
\addauthor{Yuezun Li}{liyuezun@ouc.edu.cn}{1,\blue{\text{\Letter}}}
\addauthor{Bo Peng}{bo.peng@nlpr.ia.ac.cn}{2}
\addauthor{Jiaran Zhou}{zhoujiaran@ouc.edu.cn}{1}
\addauthor{Huiyu Zhou}{hz143@leicester.ac.uk}{3}
\addauthor{Junyu Dong}{dongjunyu@ouc.edu.cn}{1}

\addinstitution{
    School of Computer Science and Technology,  \\
    Ocean University of China\\
}
\addinstitution{
 New Laboratory of Pattern Recognition (NLPR),  \\
 Institute of Automation, Chinese Academy of Sciences (CASIA)
}
\addinstitution{
 School of Computing and Mathematical Sciences, \\
 University of Leicester \\
 \vspace{0.5cm}
 \blue{\text{\Letter}}: \footnotesize{Corresponding author}
}

\runninghead{Zhang et al}{Mumpy}

\usepackage{times}
\usepackage{soul}
\usepackage{url}
\usepackage{graphicx}
\usepackage{amsmath}
\usepackage{amsthm}
\usepackage{amssymb}
\usepackage{multirow}
\usepackage{booktabs}
\usepackage{algorithm}
\usepackage{algorithmic}
\usepackage{bm}
\usepackage[switch]{lineno}
\usepackage{float}
\usepackage{subfigure}
\usepackage{wrapfig}
\usepackage{caption}

\usepackage{float}
\usepackage{colortbl}

\def\eg{\emph{e.g.}}

\def\ie{\emph{i.e.}}

\definecolor{hl}{gray}{.9}

\def\blue#1{\textcolor{blue}{#1}}

\def\eg{\emph{e.g}\bmvaOneDot}

\begin{document}

\maketitle

\begin{abstract}
The task of video inpainting detection is to expose the pixel-level inpainted regions within a video sequence. Existing methods usually focus on leveraging spatial and temporal inconsistencies. However, these methods typically employ fixed operations to combine spatial and temporal clues, limiting their applicability in different scenarios. In this paper, we introduce a novel Multilateral Temporal-view Pyramid Transformer ({\em MumPy}) that collaborates spatial-temporal clues flexibly. Our method utilizes a newly designed multilateral temporal-view encoder to extract various collaborations of spatial-temporal clues and introduces a deformable window-based temporal-view interaction module to enhance the diversity of these collaborations. Subsequently, we develop a multi-pyramid decoder to aggregate the various types of features and generate detection maps. By adjusting the contribution strength of spatial and temporal clues, our method can effectively identify inpainted regions. We validate our method on existing datasets and also introduce a new challenging and large-scale Video Inpainting dataset based on the YouTube-VOS dataset, which employs several more recent inpainting methods. The results demonstrate the superiority of our method in both in-domain and cross-domain evaluation scenarios.
\end{abstract}

\section{Introduction}
\label{sec:intro}
Video inpainting is an emerging technique that aims to recover the disrupted or designated regions in sequences while maintaining consistent semantic context \cite{kim2019deep,oh2019onion,lee2019cpnet,liu2021fuseformer,ren2022dlformer,zhou2023propainter}. In recent years, there has been remarkable progress in this technique with the ever-growing deep generative methods. As a result, the realism of inpainted visuals has significantly improved, making it increasingly challenging for human observers to detect the manipulations. The misuse of this technique can lead to serious concerns, such as removing crucial evidence or objects to deceive judges or mislead public opinion \cite{li2021noise,dong2022mvss,sun2023safl}. Therefore, detecting video inpainting has become {an urgent need} in digital forensics, especially with the rapid advancement of large-scale vision generative models \cite{goodfellow2020generative,harshvardhan2020comprehensive,yang2022diffusion}.  

\begin{figure}[!t]
\centering 
\includegraphics[width=0.9\linewidth]{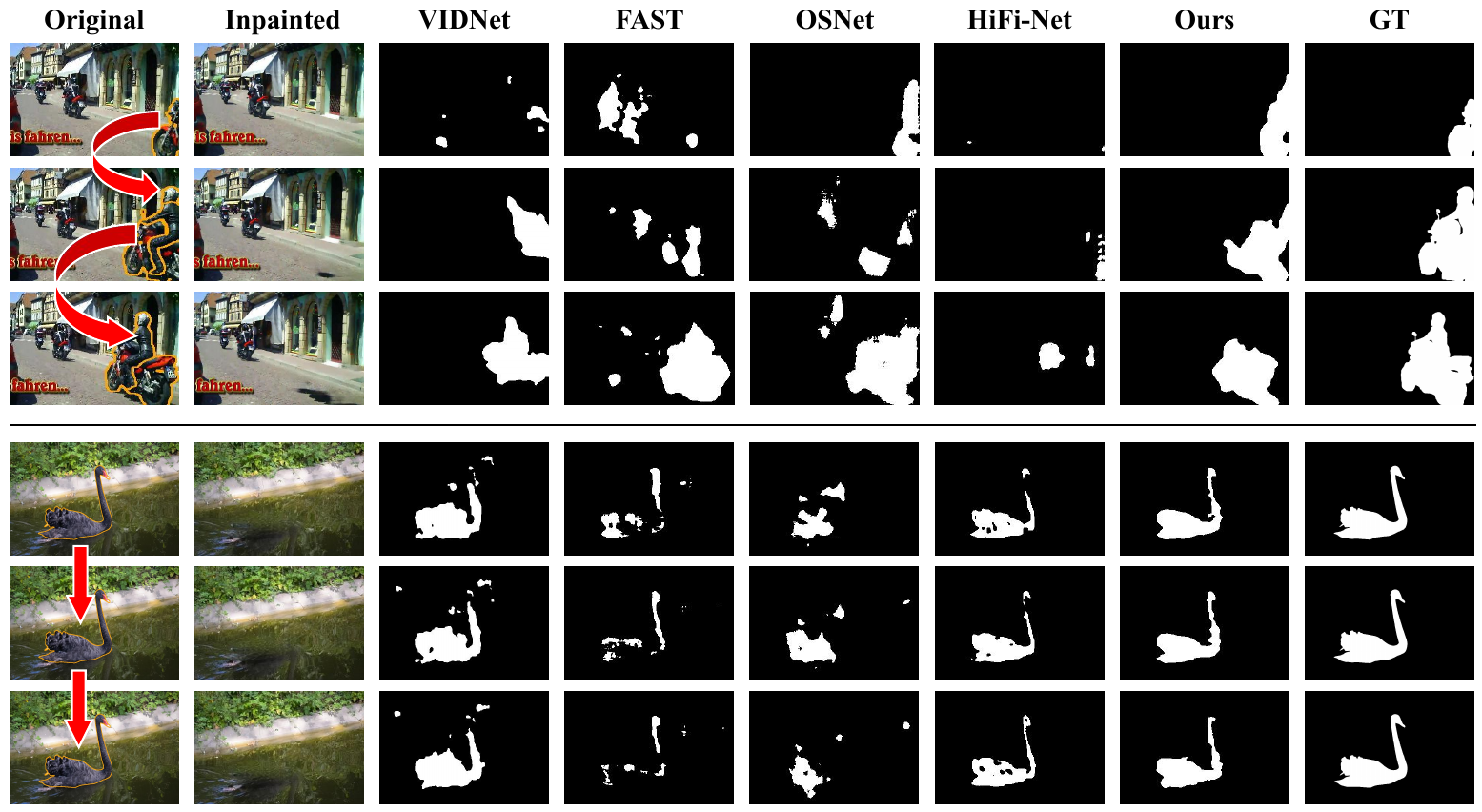}
\vspace{0.1cm}
\caption{\footnotesize 
{Results of our method compared with the others in cross-domain scenarios.}
The top examples show an obvious temporal relationship while the bottom ones exhibit a strong spatial relationship. These examples demonstrate the significance of flexible collaboration of spatial-temporal clues.} 
\label{Fig.introduction}
\vspace{-0.4cm}
\end{figure}

In recent years, many methods have been proposed for video inpainting detection \cite{zhou2021vid,yu2021frequency,wei2022deep}. One typical solution to {detect} 
the {inpainted} regions is by leveraging the frame-level spatial inconsistency clues, \eg, \cite{li2019localization,li2021noise,li2023transformer}. Due to the nature of the video sequence, inpainted regions can also introduce temporal inconsistency. Thus, more recent methods incorporate sequential-based strategies to explore both spatial and temporal inconsistencies \cite{zhou2021vid,yu2021frequency,wei2022deep}. {Typically, these methods employ deep models to extract spatial and temporal features using fixed operations (\eg, convolution with fixed kernel size) for all video sequences, leading to a straightforward combination of these features.}
However, the importance of spatial or temporal clues may differ across various scenarios. For example, in videos having a strong temporal relationship (\eg, target objects have obvious movement in sequence), the inpainted regions are more likely to disrupt temporal consistency. In such cases, paying more attention to temporal clues {allows one to identify} the manipulation. On the other hand, in videos with strong spatial relationships (\eg, target objects are significantly salient), spatial inconsistency can more effectively identify the manipulated regions. Thus, a more feasible collaboration of spatial-temporal clues is needed in practical applications (see Fig.~\ref{Fig.introduction}).


In this paper, we describe a {\em Mu}ltilateral Te{\em m}poral-view {\em Py}ramid Transformer ({\em MumPy}) to expose inpainted regions by adopting a variety of collaboration ways for spatial-temporal clues (see Fig.~\ref{Fig.main}). Our method considers different degrees of importance for spatial and temporal clues and adjusts the contribution strength of each clue accordingly. To achieve this, we develop a Multilateral Temporal-view Encoder (Sec.~\ref{sec:encoder}), a Deformable Window-based Temporal-view Interaction module (Sec.~\ref{sec:interaction}), and a Multi-pyramid Decoder (Sec.~\ref{sec:decoder}), respectively. 
The Multilateral Temporal-view Encoder consists of several branches, each representing a specific collaboration of spatial-temporal clues within different temporal views. 
To increase the diversity of collaborations, we propose a {deformable window-based temporal-view interaction} strategy that fuses the knowledge from adjacent branches. The motivation is a single branch can represent a more comprehensive temporal view if it absorbs the knowledge from others. This strategy adaptively builds the correlation between inpainted regions from different temporal views using deformable-style attention and performs the interaction separately inside windows. 
Furthermore, we propose a multi-pyramid decoder to generate the detection maps. This decoder makes full use of the intermediate features from the encoder, the frequency signals, and the fusing of high-level features in multiple pyramid schemes. 

Following previous works \cite{zhou2021vid,yu2021frequency,wei2022deep}, we validate our method on the DAVIS Video Inpainting dataset (DVI) and Free-from Video Inpainting dataset (FVI), demonstrating its efficacy in both in-domain and cross-domain evaluation scenarios. To provide a more comprehensive evaluation, we present a new challenging and large-scale video inpainting dataset based on YouTube-VOS \cite{xu2018youtube}. Compared to DVI ({$150$ videos) and FVI ($100$ videos), this dataset contains more videos ({$3471$ in total}) covering diverse real-world scenarios and incorporates four additional state-of-the-art video inpainting methods (Sec.~\ref{sec:YTVI}). Our method is further evaluated on this dataset following the same evaluation scenarios, showing its superiority compared to recent counterparts.

Our contributions can be summarized into three-fold: \textbf{1)} We describe a new network, {\em MumPy}, that allows for flexible exploration of spatial-temporal correlations, and propose a novel temporal-view interaction strategy to enhance the diversity of the collaborations between spatial and temporal clues. \textbf{2)} We introduce a multi-pyramid decoder that effectively leverages knowledge from various sources, including the accumulation of features from the encoder, the assistance of frequency signals, and the guidance of high-level features. \textbf{3)} We present a challenging and large-scale YouTube-VOS Video Inpainting dataset (YTVI), which includes many state-of-the-art video inpainting methods. Extensive experiments on the YTVI dataset, as well as the DVI and FVI datasets, demonstrate the superior performance of our method in both in-domain and cross-domain evaluation.

\begin{figure*}[!t]
\centering 
\includegraphics[width=0.9\textwidth]{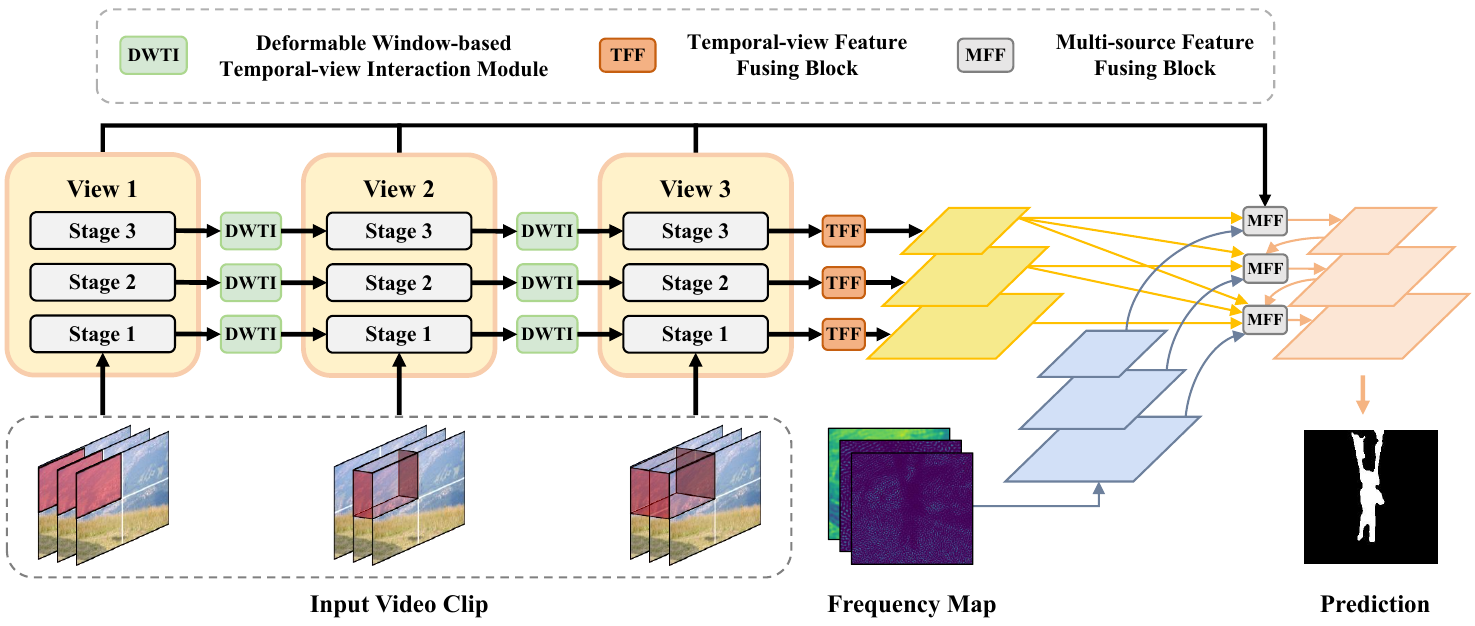}
\vspace{0.1cm}
\caption{\footnotesize Overview of the proposed Multilateral Temporal-view Pyramid Transformer. See text for details.} 
\label{Fig.main}
\vspace{-0.3cm}
\end{figure*}

\vspace{-0.4cm}
\section{Related Work}
\vspace{-0.2cm}
\smallskip
\noindent{\bf Video Inpainting.}
The improvement of deep generative techniques has greatly made video inpainting effortless and realistic. Notably, recent inpainting methods such as VI \cite{kim2019deep}, OP \cite{oh2019onion}, and CP \cite{lee2019cpnet} have drawn a lot of attention. They are involved in DVI dataset due to their favorable performance. More recently, several methods have been proposed to further improve the quality of inpainting using more advanced deep models. For instance, ISVI \cite{Zhang_2022_CVPR} employs optical flows for inpainting propagation, while EG2 \cite{liCvpr22vInpainting} performs propagation in feature space. FF \cite{liu2021fuseformer} introduces a Transformer model to explore fine-grained information for inpainting, and PP \cite{zhou2023propainter} incorporates optical flow with Transformers for inpainting.

\smallskip
\noindent{\bf Video Inpainting Detection.} To detect inpainting videos, many methods have been proposed, \eg, \cite{li2019localization,li2021noise,li2023transformer,zhou2021vid,yu2021frequency}. One direction of the methods focuses on capturing the frame-level inpainting clues. For instance, \cite{li2019localization} proposes high-pass pre-filtering to acquire high-frequency residual information, assisting in locating {inpainted} regions. \cite{li2021noise} exploits existing noise discrepancies between authentic and inpainted images. Since these methods focus on the spatial traces at the frame level, they could not utilize the temporal information exhibited in inpainted videos, which possibly degrades their performance in real-world scenarios with prominent object motion. 

To leverage the temporal information, there are several methods proposed to take video clips as input, \eg, \cite{zhou2021vid,wei2022deep,yu2021frequency}. The works of \cite{zhou2021vid,wei2022deep} extract rich spatial features and utilize LSTM-related structures to extract temporal features. The recent Video Vision Transformers greatly improve the ability to model spatial-temporal features for classification \cite{arnab2021vivit,bertasius2021space,yan2022multiview,park2023dual}. 
Inspired by these architectures, the work of \cite{yu2021frequency} introduces spatial-temporal patches obtained from extracted video clips into a vision transformer to model spatial-temporal correlations among patches. However, the existing methods have fixed collaborations between spatial and temporal features, making them less adaptable to various real-world scenarios. Therefore, this paper introduces a flexible approach that enables multiple temporal-view collaborations with dedicated architecture design.

\vspace{-0.3cm}
\section{Method}

\subsection{Multilateral Temporal-view Encoder}
\label{sec:encoder}
Denote a video clip as a set of frames ${\cal V} = [ {\cal I}_{1}, ..., {\cal I}_{T} ]$,
where ${\cal I}_{i} \in \mathbb{R}^{H \times W \times C}$ denotes the $i$-th frame. 
The general video-based Transformers first divide it into several spatial and temporal patches, known as tubelets. 
Denote these tubelets as $\{b_1, ..., b_N  \}$, where $b_i \in \mathbb{R}^{t \times h \times w \times c}$ and $N = \lfloor \frac{T}{t} \rfloor \times \lfloor \frac{H}{h} \rfloor \times \lfloor \frac{W}{w} \rfloor$. Note that in existing methods, the length of tubelets $t$ is typically fixed, indicating a fixed collaboration of spatial and temporal clues. In our method, we consider various collaboration ways of these two clues, inspired by the multimodal spirits \cite{fan2021multiscale,yan2022multiview,huang2022multi}. 
Since $t$ determines the length of the temporal window, different values correspond to different temporal views, and these views indicate various collaborations of spatial and temporal clues. For example, $t = T$ denotes to take the whole sequence into account, while $t=1$ is degraded to only use spatial patches. Thus we transform the input sequence into multiple types of tokens according to different $t$. Denote the set of temporal-view (tubelets length) as $\{t_1,...,t_m  \}$, where $t_i \leq t_{i+1}, t_i \geq 1, t_{i+1} \leq T$ and $m$ is the number of temporal-view. Given the video clip ${\cal V}$, we employ 3D convolution as in \cite{arnab2021vivit} for tokenization. For a temporal view $t \in \{ t_1,...,t_m \}$, the convolution kernel is set to the size of $t \times h \times w$ with stride $t \times h \times w$. Denote the 3D convolution operations at temporal-view $t$ as ${\cal T}^{(t)}$. \textcolor{black}{The corresponding tokens from the video clip can be defined as $\mathbf{z}^{0,(t)} = [ z^{(t)}_i, ..., z^{(t)}_N ]$, }where $z^{(t)}_i = {\cal T}^{(t)}(b_i)$. Thus we obtain a set of input tokens according to different temporal views as $\{ \mathbf{z}^{{0,}(t_i)},...,\mathbf{z}^{{0,}(t_m)} \} $.

For each temporal view, we build a $k$-stage Transformer network, and each stage contains several pairs of Swin Transformer blocks~\cite{liu2021swin}. 
The operation of consecutive Swin Transformer blocks can be defined as
\begin{equation}
\small
    \begin{aligned}
        & \mathbf{\hat{z}}^{l, (t)} = \text{W-MSA} \left ( \text{LN}(\mathbf{z}^{l-1, (t)}) \right )+ \mathbf{z}^{l-1, (t)}, \; \mathbf{z}^{l, (t)} = \text{MLP} \left ( \text{LN}(\mathbf{\hat{z}}^{l, (t)}) \right ) + \mathbf{\hat{z}}^{l, (t)}, \\
        & \mathbf{\hat{z}}^{l+1, (t)} = \text{SW-MSA} \left ( \text{LN}(\mathbf{z}^{l, (t)}) \right )+ \mathbf{z}^{l, (t)}, \; \mathbf{z}^{l+1, (t)} = \text{MLP} \left ( \text{LN}(\mathbf{\hat{z}}^{l+1, (t)}) \right ) + \mathbf{\hat{z}}^{l+1, (t)}, \\
    \end{aligned}
\end{equation}
where W-MSA and SW-MSA denote window-based multi-head self-attention and shifted window configuration, LN is layer normalization, MLP is multi-layer perception blocks and $l$ is the index of the Transformer block, respectively. Note that $l=0$ denotes the input tokenization.
After this, we obtain a set of tokens corresponding to different temporal views and merge these tokens as the input of Multi-source Feature Fusing Pyramid (see Sec.~\ref{sec:decoder}).


\subsection{Deformable Window-based Temporal-view Interaction}
\label{sec:interaction}


Ideally, each view can interact with all other views. Yet, this will incur large computational costs. Hence, we opt to perform interaction solely among adjacent branches, following an ascending order of temporal-view. The rationale behind this is that larger views concentrate more on temporal relationships, while smaller views offer richer spatial relationships. Interacting in this way can increase the diversity of larger views. 
The interaction is accomplished by the cross-attention mechanism. Note that conventional cross-attention operations build the correlations across all feature elements, which may overlook the importance of inpainted regions. To solve this, we propose Deformable Window-based Temporal-view Interaction (DWTI), which employ the deformable attention mechanism \cite{xia2022vision} to adaptively concentrate on the inpainted regions. Fig.~\ref{Fig:interaction} illustrates an overview of this process.
Denote two intermediate tokens from adjacent temporal views as $\mathbf{z}^{(t)}$ and $\mathbf{z}^{(t+1)}$\footnote{We omit the index of Transformer block $l$ for simplicity.}, where the former comes from a smaller view and the other from a larger view. We convert these two tokens to the same shape of $h' \times w' \times c'$, \ie, $\mathbf{z}^{(t)} \in \mathbb{R}^{h' \times w' \times c'}, \mathbf{z}^{(t+1)} \in \mathbb{R}^{h' \times w' \times c'}$. {We} generate queries from $\mathbf{z}^{(t+1)}$ and keys, values from $\mathbf{z}^{(t)}$. The queries $q$ are obtained by linearly projecting the tokens $\mathbf{z}^{(t+1)}_i$ as $q=\mathbf{z}^{(t+1)}_i W_{q}$. Denote $p$ as the uniform grid of points. To obtain the offset for $p$, we utilize a DNN which can transform the queries $q$ into offset values as $\triangle p = \theta\left ( q \right )$.
We can obtain the deformed points by integrating reference points $p$ and corresponding offsets $\triangle p$, and sample the features at the locations of deformed points in $\mathbf{z}^{(t)}$ to project to the key and value tokens, defined as
\begin{equation}
\small
\begin{aligned}
    & q=\mathbf{z}^{(t+1)}W_{q}, k=\tilde{\mathbf{z}}^{(t)}W_{k}, v=\tilde{\mathbf{z}}^{(t)}W_{v}, \\
    & \triangle p=\theta \left ( q \right ) ,\tilde{\mathbf{z}}^{(t)}=\delta ( \mathbf{z}^{(t)};p+ \triangle  p ),
\end{aligned}
\label{eq:deformable}
\end{equation}
where $\delta(\cdot)$ is a sampling function. Then we perform MSA by using the obtained $q,k,v$ respectively as $h = \text{Softmax}(qk^{\top}/\sqrt{d})v$.

Considering that the global attention is {easy to be} disrupted by outliers and requires large computational costs{, we} describe a window-based method by only considering the cross-attention inside corresponding windows between two temporal views. {Note that each token is an integration of $K$ windows, defined as $\mathbf{z}^{(t)} = [\mathbf{z}^{(t)}_1,...,\mathbf{z}^{(t)}_K]$.} For each window, we perform the deformable cross-attention using Eq.~\eqref{eq:deformable} and then ensemble up the results.


\subsection{Multi-pyramid Decoder}
\label{sec:decoder}
We design a CNN-based decoder architecture to take in the features from the encoder and then generate the detection maps. Inspired by the pyramid spirits in computer vision tasks \cite{zhang2020feature,wang2021pyramid}, we propose a Multi-pyramid architecture to fully utilize different features. 


\begin{figure}[!t]
    \centering
    \begin{minipage}[t]{0.36\linewidth}
    \centering 
    \includegraphics[width=\linewidth]{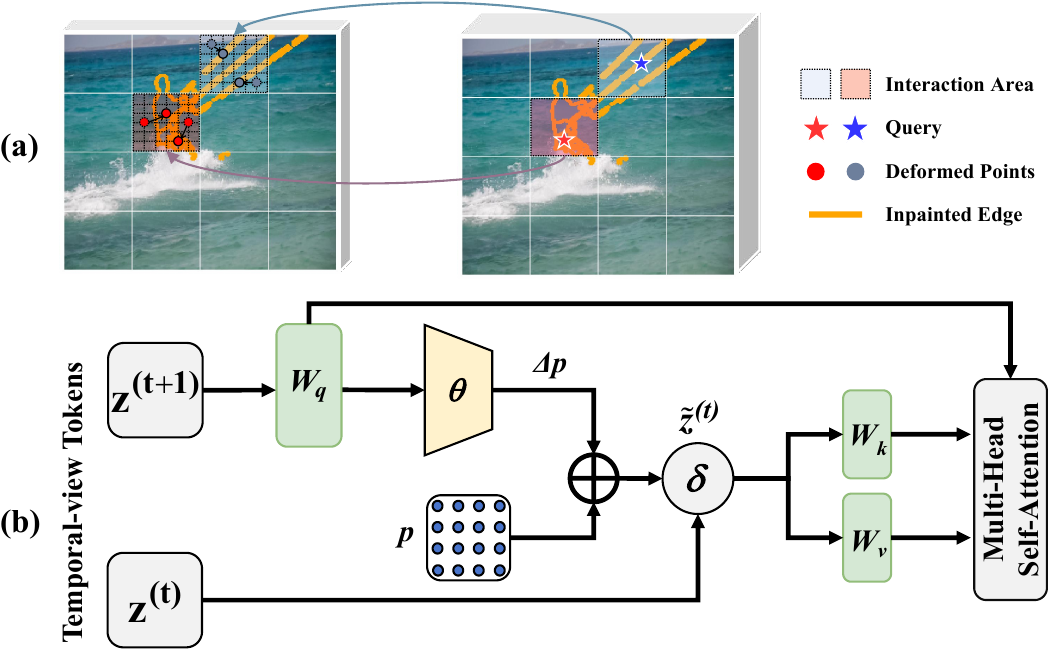}
    \vspace{-0.2cm}
    \caption{\footnotesize (a) Diagram and (b) process of DWTI. } 
    \label{Fig:interaction}
    \end{minipage}
    \hfill
    \begin{minipage}[t]{0.25\linewidth}
    \centering
    \includegraphics[width=1\linewidth]{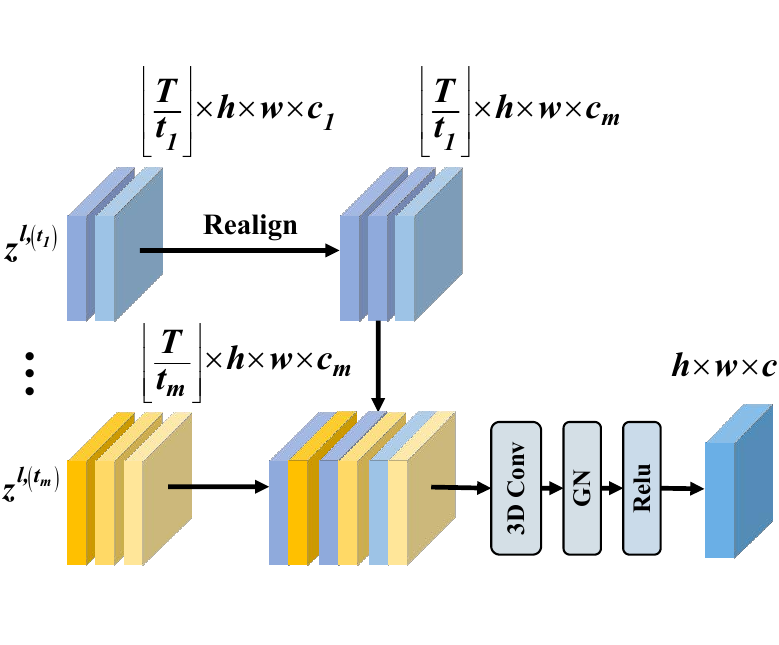}
    \vspace{-0.2cm}
    \caption{\footnotesize Temporal-view Feature Fusing (TFF) block. }
    \label{Fig:tff}
    \end{minipage}
    \hfill
    \begin{minipage}[t]{0.36\linewidth}
    \centering
    \includegraphics[width=1\linewidth]{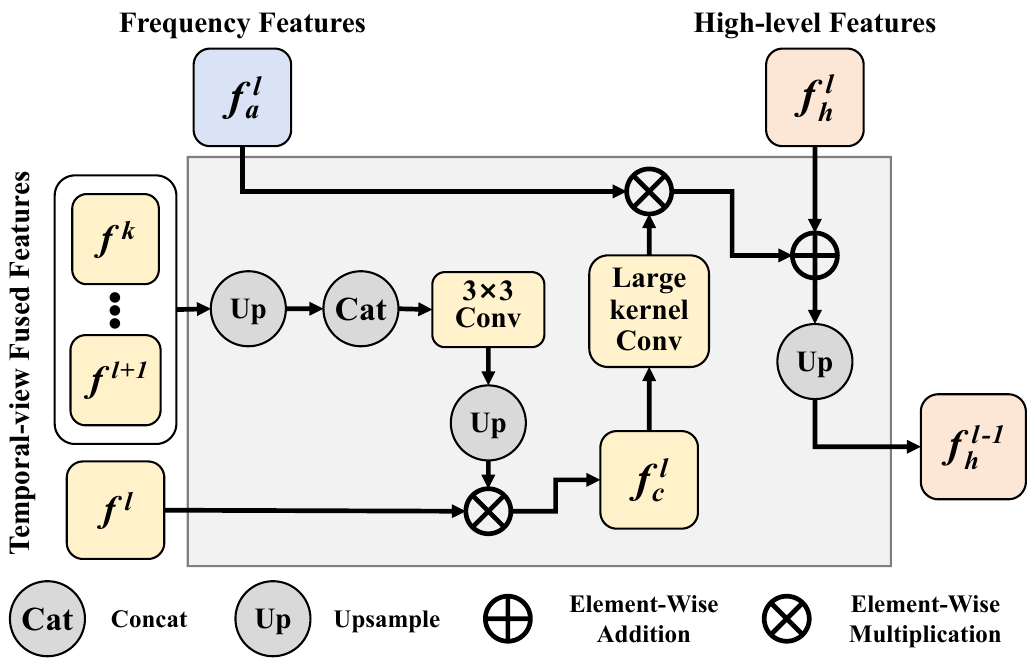}
    \vspace{-0.2cm}
    \caption{\footnotesize Multi-source Feature Fusing (MFF) block. } 
    \label{Fig:fab}
    \end{minipage}
\end{figure}


\smallskip
\noindent{\bf {Temporal-view Pyramid.}}
The final features from the encoder likely represent high-level information of inpainting traces but lack detailed information, which impedes their detection ability when applied in complicated scenarios. Therefore, we integrate the intermediate features of different temporal views into the decoder to provide more instructive guidance. 

Firstly, we develop a {\em Temporal-view Feature Fusing Block} (TFF) to fuse features from different views at the same stage. Denote $\mathbf{z}^{l,(t_1)},...,\mathbf{z}^{l,(t_m)}$ as the intermediate features of the temporal-view from $t_1$ to $t_m$ at $l$-th stage, where $\mathbf{z}^{l,(t_1)} \in \mathbb{R}^{\lfloor \frac{T}{t_1} \rfloor \times h \times w \times c_1},...,\mathbf{z}^{l,(t_m)} \in \mathbb{R}^{\lfloor \frac{T}{t_m} \rfloor \times h \times w \times c_m}$ respectively. Denote $v$ as the maximum of $\{ \lfloor \frac{T}{t_1} \rfloor,...,\lfloor \frac{T}{t_m} \rfloor\}$. To fuse these features, this block first realigns their dimensions by expanding the first dimension to $v$. Taking $\mathbf{z}^{l,(t_1)}$ for instance, its size is aligned as ${\lfloor \frac{T}{t_1} \rfloor \times h \times w \times c_1} \rightarrow {v \times h \times w \times c_1}$. Then we concatenate these features along the channel dimension and send them into a 3D convolution layer, which is followed by a Group Normalization to generate the fuse feature $\bm{f}^{l} \in \mathbb{R}^{h \times w \times c}$. For $k$ stages in our method, we obtain a set of fused features $\{ \bm{f}^{1}, ..., \bm{f}^{k} \}$. These operations are shown in Fig.~\ref{Fig:tff}.

\smallskip
\noindent{\bf {Frequency-assistance Pyramid.}} Inspired by the successful applications of frequency features in forensics \cite{qian2020thinking,yu2021frequency,gu2022exploiting}, we extract the frequency features as an auxiliary to enhance detection. Given the video clip $\mathcal{V} = [ \mathcal{I}_1, ..., \mathcal{I}_{T} ]$. We use the middle frame as a target to extract frequency features. We first employ Discrete Cosine Transform (DCT) \cite{ahmed1974discrete} to convert this frame into frequency maps. Then we employ three frequency band-pass filters to decompose the frequency map into low-pass, mid-pass, and high-pass signals \cite{qian2020thinking}. Based on these signals, we employ Inversed Discrete Cosine Transform (IDCT) to transform each signal as a frequency feature and concatenate all the transformed features along the channel dimension as the final frequency features, denoted as $\bm{f}^{0}_{a} \in \mathbb{R}^{H \times W \times 3C}$. Then we build a frequency pyramid by adopting AvgPooling to reduce the size progressively as $\{ \bm{f}^{1}_a, ..., \bm{f}^{k}_a \}$.  



\smallskip
\noindent{\bf {Multi-source Feature Fusing Pyramid.}}
We propose a {\em Multi-source Feature Fusing Block} (MFF) to fuse the temporal-view features, frequency features, and high-level features in a pyramid manner (see Fig.~\ref{Fig:fab}). 
For temporal-view features, at the $l$-th stage, we upsample and concatenate all deeper features and send them into a convolution layer. Then the dot-production is performed with the feature of the current stage. Performing dot-production instead of summation can instruct the current feature to focus on more important spots represented by the deep accumulated features. After these operations, we obtain accumulated features of $k$ stages as $\{ \bm{f}^{1}_c, ..., \bm{f}^{k}_c \}$.
We then fuse frequency features in an attention fashion.
Note that $\bm{f}^{l}_a$ has the same height and width with $\bm{f}^{l}_c$. For each stage, we fuse the semantic features $\bm{f}^{l}_c$ with the frequency feature $\bm{f}^{l}_a$ as follows. We first perform a large kernel convolution \cite{peng2017large} on $\bm{f}^{l}_c$ and then employ dot-production with $\bm{f}^{l}_a$. Denote the features after these operations as $\{ \bm{f}^{1}_{c'}, ..., \bm{f}^{k}_{c'} \}$. The motivation is that the frequency feature can instruct the semantic feature paying more attention to the frequency-importance spots. 

Since the high-level features can provide important guidance, we fuse them with other features and convey them to the next pyramid layer. Denote the output features from the encoder as $\bm{f}_{h}$. They are integrated into the last obtained feature of $k$-th stage $\bm{f}^{k}_{c'}$ as $\bm{f}^{k}_{h}$ to provide high-level guidance. Then $\bm{f}^{k}_{h}$ is upsampled and sent into the next pyramid layer. The integration process is iterative, which builds another pyramid structure. 

\subsection{Objective Functions}
Our objectives consist of two components. The first component measures the loss of mean Intersection-over-Union (mIoU) between the predicted detection mask $\mathcal{M}$ and the ground truth mask $\mathcal{M}_{\rm GT}$, which can be written as
\begin{equation}
\small
\begin{aligned}
    \mathcal{L}_{1}(\mathcal{M}, \mathcal{M}_{\rm GT}) = 1 - \frac{\sum \left ( \mathcal{M} \cdot \mathcal{M}_{\rm GT} \right )}{\sum \left ( \mathcal{M} + \mathcal{M}_{\rm GT} - \mathcal{M} \cdot \mathcal{M}_{\rm GT} \right )}.
\end{aligned}
\end{equation}
The second component is a focal cross-entropy loss. Observing that the inpainted region is usually less than the authentic region. Thus we use focal loss here to mitigate this imbalance. This loss term can be defined as
\begin{equation}
\small
\begin{aligned}
    \mathcal{L}_{2}(\mathcal{M}, \mathcal{M}_{\rm GT}) = & - \sum \left ( \alpha \cdot (1 - \mathcal{M})^{\gamma} \cdot \mathcal{M}_{\rm GT} \log (\mathcal{M}) \right. \\
    & \left.  + (1 - \alpha) \mathcal{M}^{\gamma} (1 - \mathcal{M}_{\rm GT}) \log (1 - \mathcal{M}) \right ),
\end{aligned}
\end{equation}
where $\alpha$ is the weight parameter to balance inpainted and authentic pixels and $\gamma$ is the parameter for hard mining. The overall objectives can be expressed as $\mathcal{L}(\mathcal{M}, \mathcal{M}_{\rm GT}) = \lambda_{1} \mathcal{L}_{1}(\mathcal{M}, \mathcal{M}_{\rm GT}) + \lambda_2 \mathcal{L}_{2}(\mathcal{M}, \mathcal{M}_{\rm GT})$, where $\lambda_{1},\lambda_{2}$ are the weight parameters to balance the losses.

\section{Youtube-vos Video Inpainting Dataset}
\label{sec:YTVI}
Davis Video Inpainting dataset (DVI) and the Free-from Video Inpainting dataset (FVI) are two existing datasets widely used in previous works \cite{zhou2021vid,yu2021frequency,wei2022deep}.
DVI is constructed on DAVIS 2016 \cite{Perazzi2016} using inpainting methods VI~\cite{kim2019deep}, OP~\cite{oh2019onion} and CP~\cite{lee2019cpnet} respectively. Each inpainting method corresponds to $50$ videos. 
FVI dataset contains 100 test videos that are processed by object removal, and are usually used for demonstrating detection generalization~\cite{chang2019free}. 
In this paper, we introduce a more challenging and large-scale {\em Youtube-vos Video Inpainting dataset (YTVI)} for a more comprehensive assessment. This dataset is built upon Youtube-vos 2018 \cite{xu2018youtube}, which contains $3471$ videos with $5945$ object instances in its training set. Since only the training set of this dataset is fully annotated, we use it to construct YTVI. Specifically, with the goal of further improving the comprehensiveness, we adopt many more recent video inpainting methods on this dataset, including EG2~\cite{liCvpr22vInpainting}, FF~\cite{liu2021fuseformer}, PP~\cite{zhou2023propainter}, and ISVI~\cite{Zhang_2022_CVPR}, together with VI, OP and CP. These inpainting methods are applied to the object regions annotated by ground truth masks.

\section{Experiments}
\smallskip
\noindent{\bf Datasets and Metrics.}
Our method is evaluated on the DVI and FVI datasets, as well as the newly proposed YTVI dataset. The detection performance is evaluated using the mean Intersection-over-Union (mIoU) and F1 score as in previous works. For clarification, the threshold used in mIoU and F1 score is set to $0.5$ for a fair comparison.


\smallskip
\noindent{\bf Implementation Details.}
Our method is implemented using PyTorch \cite{paszke2019pytorch} with a GeForce RTX 3090Ti. For the backbone of the multilateral temporal-view encoder, we develop three branches and employ the Tiny, Small, and Base variants of the Swin Transformer in each branch. We use ViT Base architecture for the global encoder. The input size of video clips is $224 \times 224$ with the sequential length of $3$ and is augmented by various common operations. In the training phase, we set the batch size to $12$, and employ an SGD optimizer with a learning rate of $0.001$ for the encoder and $0.01$ for the decoder. We set the weight decay to $10^{-4}$ and use the poly learning rate decay to adjust the learning rate from the initialization to $10^{-5}$. 


\smallskip
\noindent{\bf Results on YTVI and DVI.}
Table~\ref{tab:YTVIcomparision} shows the performance of different detection methods on the YTVI dataset under both In-inpainting and Cross-inpainting evaluation. These methods are trained on two inpainting methods (marked *) and tested on all methods. For example, (VI*, OP*, CP) indicates that the method is trained on videos manipulated by VI and OP, and then tested on VI, OP, and CP respectively. It can be seen that our method outperforms others by a large margin. For example, in the setting of (VI*, OP*, CP), our method improves around $10.5\%$ (mIoU) and $11.5\%$ (F1) under {averaged} In-inpainting evaluation, and $7\%$ (mIoU) and $9\%$ (F1) under Cross-inpainting evaluation, compared to the second-best method OSNet. A similar trend is also observed in the other training settings (Right two groups in Table \ref{tab:YTVIcomparision}). Since the YTVI dataset is composed of real-world videos with high diversity, the proposed flexible collaboration ways of spatial-temporal clues exhibit significant advantages in comparison with the existing counterparts. 
Table~\ref{tab:DVIcomparision} shows the performance of different methods on DVI dataset, revealing a similar trend as in Table \ref{tab:YTVIcomparision}. 

\begin{table*}[!t]
    \small
    \centering
    \caption{\footnotesize {Performance of different methods on YTVI dataset}. (a*,b*,c) denotes each method is trained on two inpainting methods (a,b), and tested on all inpainting methods (a,b,c).}
    \vspace{0.2cm}
    \resizebox{\linewidth}{!}{
    \begin{tabular}{l|ccc|ccc|ccc}
        \hline
         \multirow{2}{*}{\textbf{Methods}} & \multicolumn{1}{c}{\textbf{VI}*}  & \multicolumn{1}{c}{\textbf{OP}*}  & \multicolumn{1}{c|}{\textbf{CP}}   & \multicolumn{1}{c}{\textbf{VI}}  & \multicolumn{1}{c}{\textbf{OP}*}  & \multicolumn{1}{c|}{\textbf{CP}*}   & \multicolumn{1}{c}{\textbf{VI}*}  & \multicolumn{1}{c}{\textbf{OP}}  & \multicolumn{1}{c}{\textbf{CP}*}   \\
         & {mIoU/F1}& {mIoU/F1}& {mIoU/F1} & {mIoU/F1}& {mIoU/F1}& {mIoU/F1} & {mIoU/F1}& {mIoU/F1}& {mIoU/F1}\\ 
         \hline
        HPF (ICCV'19) \cite{li2019localization} &0.50/0.63 &0.48/0.59 &0.42/0.54 &0.12/0.18 &0.47/0.59 &0.52/0.64 & 0.52/0.65 &0.13/0.20 &0.55/0.69 \\
        GSRNet (AAAI'20) \cite{zhou2020generate}& 0.51/0.64  &0.50/0.63 &0.38/0.50 &0.14/0.22 &0.51/0.63&0.62/0.73 & 0.49/0.62 &0.21/0.32 &0.55/0.68 \\
        VIDNet (BMVC'21) \cite{zhou2021vid}&0.62/0.74 &0.51/0.64  &0.43/0.56 &0.15/0.23 &0.54 /0.66 &0.62/0.73 &0.62/0.74 &0.20/0.28 &0.61/0.72 \\
        FAST (ICCV'21) \cite{yu2021frequency}& 0.49/0.61 & 0.54/0.66 & 0.46/0.58 & 0.25/0.35 & 0.47/0.58 & 0.60/0.71 & 0.47/0.59 & 0.29/0.40 & 0.62/0.73\\
        OSNet (CVPR'22) \cite{wu2022robust}& 0.60/0.70 & 0.58/0.67 & 0.56/0.66 & 0.17/0.23 & 0.61/0.71 & 0.69/0.78 & 0.65/0.74 &0.30/0.40 &0.70/0.78 \\
        HiFi-Net (CVPR'23)\cite{guo2023hierarchical}&0.64/0.74 & 0.35/0.46 & 0.39/0.50 & 0.17/0.24 & 0.34/0.44 & 0.57/0.67 & 0.62/0.73 & 0.08/0.11 & 0.50/0.60\\
        IML-ViT (AAAI'24) \cite{ma2023iml} & 0.60/0.72 & 0.56/0.68 & 0.55/0.67 & 0.22/0.32 & 0.60/0.71 & 0.69/0.79 & 0.60/0.71 & 0.25/0.35 & 0.66/0.76\\
        \rowcolor{hl} Ours & \textbf{0.72/0.82} & \textbf{0.67/0.78} & \textbf{0.63/0.75} & \textbf{0.33/0.45} & \textbf{0.70/0.79} & \textbf{0.75/0.84} & \textbf{0.73/0.82} & \textbf{0.42/0.54} & \textbf{0.73/0.83}\\
        \hline
    \end{tabular}
    }
    \label{tab:YTVIcomparision}
\end{table*}

\begin{table*}[!t]
    \small
    \centering
    \caption{\footnotesize {Performance of different methods on DVI dataset.} }
    \vspace{0.2cm}
    \resizebox{\linewidth}{!}{
    \begin{tabular}{l|ccc|ccc|ccc}
        \hline
         \multirow{2}{*}{\textbf{Methods}} & \multicolumn{1}{c}{\textbf{VI}*}  & \multicolumn{1}{c}{\textbf{OP}*}  & \multicolumn{1}{c|}{\textbf{CP}}   & \multicolumn{1}{c}{\textbf{VI}}  & \multicolumn{1}{c}{\textbf{OP}*}  & \multicolumn{1}{c|}{\textbf{CP}*}   & \multicolumn{1}{c}{\textbf{VI}*}  & \multicolumn{1}{c}{\textbf{OP}}  & \multicolumn{1}{c}{\textbf{CP}*}   \\
         & {mIoU/F1}& {mIoU/F1}& {mIoU/F1} & {mIoU/F1}& {mIoU/F1}& {mIoU/F1} & {mIoU/F1}& {mIoU/F1}& {mIoU/F1}\\ 
         
         \hline
        HPF (ICCV'19) \cite{li2019localization} &0.46/0.57 &0.49/0.62 &0.46/0.58 &0.34/0.44 &0.41 /0.51 &0.68/0.77 & 0.55/0.67 &0.19/ 0.29 &0.69/0.80 \\
        GSRNet (AAAI'20) \cite{zhou2020generate} & 0.57/0.69  &0.50/0.63 &0.51/0.63 &0.30 /0.43 &0.74/0.82&0.80/0.85 & 0.59 /0.70 &0.22/0.33 &0.70/0.77 \\
        VIDNet (BMVC'21) \cite{zhou2021vid} &0.59/0.70 &0.59/0.71  &0.57/0.69 &0.39/0.49 &0.74/0.82 &0.81/0.87 &0.59/0.71 &0.25/0.34 &0.76/0.85 \\
        FAST (ICCV'21) \cite{yu2021frequency} & 0.61/0.73 & 0.65/0.78 & 0.63/0.76 & 0.32/0.49 & 0.78/0.87 & 0.82/0.90 & 0.57/ 0.68 & 0.22/0.34 & 0.76/0.83 \\
        DSTT (ICASSP'22) \cite{wei2022deep} & 0.60/0.73 & 0.69/0.80 & 0.65/0.77 & - & - & - & - & - & - \\
        OSNet (CVPR'22) \cite{wu2022robust}& 0.64/0.76 & 0.49/0.63 & 0.60/0.73 & 0.63/0.75 & 0.54/0.68 & 0.65/0.77 & 0.68/0.79 &0.36/0.50 &0.65/0.77 \\
        HiFi-Net (CVPR'23) \cite{guo2023hierarchical}& 0.71/0.81 & 0.80/0.88 & \textbf{0.74/0.83} & 0.65/0.76 & 0.80/0.88 &0.83/0.90 & 0.72/0.82 & 0.42/0.54 & \textbf{0.83/0.90}\\
        IML-ViT (AAAI'24) \cite{ma2023iml}& \textbf{0.75/0.84} & 0.69/0.80 & 0.71/0.82 & 0.68/0.80 & 0.69/0.80 &0.75/0.85 & \textbf{0.75/0.84} & 0.52/0.66 & 0.75/0.84\\
        \rowcolor{hl} Ours & 0.71/0.81 & \textbf{0.80/0.88} & 0.70/0.82 & \textbf{0.69/0.80} & \textbf{0.82/0.89} & \textbf{0.83/0.90} & 0.72/0.82 & \textbf{0.65/0.76} & 0.81/0.89\\
        \hline
        \end{tabular}
    }
    \label{tab:DVIcomparision}
\end{table*}

\begin{figure}[!t]
\small
    \centering
    \begin{minipage}[b]{0.44\textwidth}
     \centering
     \captionof{table}{\footnotesize {Cross-dataset performance of different methods from DVI to FVI dataset (DVI $\rightarrow$ FVI)}.}     
    \vspace{0.2cm}
    \resizebox{0.85\linewidth}{!}{
    \begin{tabular}{p{3cm}|c|c}
        \hline
         \multirow{2}{*}{{\textbf{Methods}}} & \multirow{2}{*}{\textbf{DVI}} &  \textbf{FVI}   \\  & & mIoU/F1  \\
         \hline
        HPF \cite{li2019localization} & \multirow{7}{*}{VI+OP} & 0.20/0.28 \\
        GSR-Net \cite{zhou2020generate}& &0.19/0.28 \\
        VIDNet \cite{zhou2021vid}& &0.25/0.36 \\
        FAST \cite{yu2021frequency}& &0.28/0.35 \\
        OSNet \cite{wu2022robust}& & 0.26/0.38\\
        HiFi-Net \cite{guo2023hierarchical}& & 0.13/0.19\\
        IML-ViT \cite{ma2023iml} & & 0.29/0.42\\
        \rowcolor{hl}Ours & &\textbf{0.36/0.48} \\
        \hline
    \end{tabular}}
    \label{tab:fvi}
    \end{minipage}
    \begin{minipage}[b]{0.5\textwidth}
    \centering
    \captionof{table}{\footnotesize {Cross-dataset performance of different methods from YTVI to DVI dataset (YTVI $\rightarrow$ DVI)}. }
    \vspace{0.2cm}
    \resizebox{0.9\linewidth}{!}{
    \begin{tabular}{l|c|ccc}
        \hline
         \multirow{3}{*}{{\textbf{Methods}}} & \multirow{3}{*}{{\textbf{YTVI}}}  & \multicolumn{3}{c}{\textbf{DVI}} \\
         & & \multicolumn{1}{c}{\textbf{VI}}  & \multicolumn{1}{c}{\textbf{OP}}  & \multicolumn{1}{c}{\textbf{CP}} \\
         & & {mIoU/F1}& {mIoU/F1}& {mIoU/F1}\\ 
         \hline
        HPF \cite{li2019localization}&\multirow{7}{*}{{VI+OP}}& 0.22/0.32  &0.37/0.49 &0.40/0.52 \\
        GSRNet \cite{zhou2020generate}& & 0.17/0.26  &0.43/0.56 &0.43/0.56 \\
        VIDNet \cite{zhou2021vid}& & 0.17/0.27  &0.29/0.41 &0.29/0.42 \\
        FAST \cite{yu2021frequency}& & 0.38/0.50  &0.59/0.71 &0.53/0.67 \\
        OSNet \cite{wu2022robust}& & 0.35/0.44  &0.49/0.61 &0.50/0.62 \\
        HiFi-Net \cite{guo2023hierarchical}& & 0.0003/0.0006  &0.65/0.76 &0.39/0.49 \\
        IML-ViT \cite{ma2023iml}& & 0.37/0.48  &0.53/0.66 &0.41/0.52 \\
        \rowcolor{hl}Ours & & \textbf{0.44/0.55}  &\textbf{0.69/0.80} &\textbf{0.64/0.76} \\
        \hline
        \end{tabular}
    }
    \label{tab:CrossDatasetevaluation}
    \end{minipage}
\end{figure}

\begin{table*}[!t]
    \small
    \centering
    \caption{\footnotesize {Cross-dataset Cross-inpainting Performance of different methods from YTVI to DVI (YTVI $\rightarrow$ DVI)}.}
    \vspace{0.2cm}
     \resizebox{\linewidth}{!}{
    \begin{tabular}{l|ccccccc|ccc}
        \hline
         \multirow{3}{*}{\textbf{Methods}} & \multicolumn{7}{c|}{\textbf{YTVI}} & \multicolumn{3}{c}{\textbf{DVI}} \\
         & \multicolumn{1}{c}{\textbf{FF}*}  & \multicolumn{1}{c}{\textbf{EG2}*}  & \multicolumn{1}{c}{\textbf{PP}*} & \multicolumn{1}{c}{\textbf{IS}}
         & \multicolumn{1}{c}{\textbf{VI}}  & \multicolumn{1}{c}{\textbf{OP}}   & \multicolumn{1}{c|}{\textbf{CP}}
         & \multicolumn{1}{c}{\textbf{VI}}  & \multicolumn{1}{c}{\textbf{OP}}   & \multicolumn{1}{c}{\textbf{CP}}\\
         & {mIoU/F1}& {mIoU/F1}& {mIoU/F1} & {mIoU/F1}& {mIoU/F1}& {mIoU/F1} & {mIoU/F1}& {mIoU/F1}& {mIoU/F1} & {mIoU/F1}\\ 
         \hline
         HPF \cite{li2019localization} & 0.47/0.58 &0.41/0.52 &0.39/0.50 &0.22/0.33 &0.14/0.22 &0.11/0.18 &0.20/0.30 &0.28/0.40 &0.12/0.19 &0.39/0.52 \\
        GSRNet \cite{zhou2020generate} & 0.71/0.81  &0.65/0.77 &0.60/0.72 &0.30/0.44 &0.06/0.11 & 0.15/0.24 & 0.19/0.29 &0.61/0.73 &0.31/0.44 &0.66/0.78 \\
        VIDNet \cite{zhou2021vid} & 0.56/0.68 &0.50/0.63  &0.47/0.59 &0.23/0.34 &0.13/0.20 &0.16/0.24 &0.31/0.43 &0.37/0.50 &0.20/0.28 &0.37/0.49 \\
        FAST \cite{yu2021frequency} & 0.54/0.66 &0.52/0.63  &0.48/0.60 &0.30/0.43 &0.19/0.29 &0.26/0.36 &0.37/0.49 &0.52/0.65 &0.40/0.52 &0.55/0.68 \\
        OSNet \cite{wu2022robust} & 0.74/0.82 &0.64/0.74  & 0.69/0.78 &0.42/0.54 &0.20/0.29 &0.30/0.39 &0.43/0.54 &0.65/0.77 &0.49/0.62 &0.65/0.77 \\
        HiFi-Net \cite{guo2023hierarchical} &0.50/0.62 &0.43/0.56  &0.33/0.45 &0.46/0.59 &0.15/0.23 &0.08/0.13 &0.15/0.22 &0.61/0.73 &0.49/0.61 &0.71/0.81 \\
        IML-ViT \cite{ma2023iml} & 0.71/0.81 & 0.67/0.78  & 0.64/0.75 & 0.40/0.54 &0.19/0.28 &0.35/0.46 &\textbf{0.54/0.67} &0.59/0.72 &0.50/0.63 & 0.64/0.76 \\
        \rowcolor{hl}Ours & \textbf{0.77}/\textbf{0.86} & \textbf{0.73}/\textbf{0.83} & \textbf{0.69}/\textbf{0.80} & \textbf{0.51}/\textbf{0.65} & \textbf{0.27}/\textbf{0.38} & \textbf{0.36}/\textbf{0.48} & {0.47}/{0.60} &\textbf{0.67}/\textbf{0.79} &\textbf{0.67}/\textbf{0.79} &\textbf{0.73}/\textbf{0.84}\\
        \hline
        \end{tabular}
     }
    \label{tab:NewInpaintingMethods}
\end{table*}

\smallskip
\noindent{\bf Results on DVI $\rightarrow$ FVI.}
Following previous works \cite{zhou2021vid,yu2021frequency}, we also evaluate the cross-dataset performance from DVI to FVI dataset. Table~\ref{tab:fvi} shows the performance of each method trained on VI+OP in DVI dataset and tested on FVI dataset. It can be seen that our method outperforms others by a large margin, {$\sim 6\%$ both in mIoU and F1 scores.} Note that FAST and VIDNet are video-based methods, which show better generalization performance than others. It confirms the significance of modeling the temporal dependencies. 

\smallskip
\noindent{\bf Results on YTVI $\rightarrow$ DVI.}
Table~\ref{tab:CrossDatasetevaluation} shows another cross-dataset experiment from YTVI to DVI dataset. Similarly, we use VI + OP in the YTVI dataset for training and evaluate all inpainting methods on the DVI dataset. We can observe our method outperforms all counterparts, improving $10\%$ on average in mIoU and F1 scores. 
Moreover, we perform a more challenging evaluation under cross-dataset cross-inpainting in Table~\ref{tab:NewInpaintingMethods}. Specifically, we use inpainting methods FF, EG2, and PP in the YTVI dataset for training and evaluate all inpainting methods on the YTVI and DVI datasets. The results show our method outperforms all the counterparts by a large margin, averaging $6.4\%$ and $6.2\%$ in mIoU and F1 scores {compared with the second-best IML-ViT}.

\begin{figure}[!t]
\small
    \centering
    \begin{minipage}[c]{0.5\textwidth}
     \centering
     \vspace{-0.1cm}
     \captionof{table}{\footnotesize Effect of each component.}
    \vspace{0.2cm}
    \tabcolsep=0.3em
    \resizebox{\linewidth}{!}{
    \begin{tabular}{ccccc|ccc}
        \hline
         \multirow{2}{*}{{\textbf{Base}}} &\multirow{2}{*}{{\textbf{TF}}}
         &\multirow{2}{*}{{\textbf{FF}}}
         & \multirow{2}{*}{{\textbf{MT}}}& \multirow{2}{*}{{\textbf{DWTI}}} 
         & {{\textbf{VI*}}}  & {{\textbf{OP}}}  & {{\textbf{CP*}}} \\
         & & & & &{mIoU/F1}& {mIoU/F1}& {mIoU/F1}\\
         \hline
         \checkmark & & &  &  & 0.64/0.76& 0.55/0.68&0.69/0.80 \\
         \checkmark & \checkmark & &  &  & 0.71/0.81& 0.59/0.72 & 0.79/0.87\\
         \checkmark & \checkmark & \checkmark &  & & 0.70/0.81& 0.60/0.73 & 0.80/0.88\\
         \checkmark & \checkmark & \checkmark & \checkmark &  & 0.72/0.82 & 0.63/0.75 & 0.82/0.89 \\
         \checkmark & \checkmark & \checkmark & \checkmark & \checkmark & 0.72/0.82 & 0.65/0.76 & 0.81/0.89\\
        \hline
        \end{tabular}
    }
    \label{tab:ablation}
    \end{minipage}
    \begin{minipage}[c]{0.4\textwidth}
     \centering
     \captionof{table}{\footnotesize {Temporal-view selection analysis}.}     
    \vspace{0.2cm}
    \resizebox{0.9\linewidth}{!}{
	\begin{tabular}{c|ccc}
            \hline
            \multirow{2}{*}{\textbf{Selection}} & {{\textbf{VI*}}}  & {{\textbf{OP}}}  & {{\textbf{CP*}}} \\ 
            & {mIoU/F1}& {mIoU/F1}& {mIoU/F1}  \\ \hline
            {View 1} & {0.71/0.82}  & {0.61/0.63}  & {0.80/0.88} \\
            {View 2} & {0.71/0.81}  & {0.59/0.72}  & {0.80/0.88}  \\
            View 3 & 0.71/0.81 & 0.57/0.69 & 0.81/0.88 \\ \hline
            {View 1,2} & {0.72/0.82}  & {0.62/0.74}  & {0.82/0.89} \\
            {View 1,3} & {0.72/0.82}  & {0.64/0.76}  & {0.82/0.89} \\
            View 2,3 & 0.71/0.81 & 0.59/0.71 & 0.81/0.89 \\ 
            View 1,2,3 & 0.72/0.82 & {0.65/0.76} & 0.81/0.89 \\ \hline
    \end{tabular}
    }
    \label{tab:viewSelection}
    \end{minipage}
\end{figure}

\begin{figure}[!t]
\vspace{-0.1cm}
\small
    \centering    
    \begin{minipage}[b]{0.52\textwidth}
    \centering
    \captionof{table}{\footnotesize {Complexity Analysis.} }
    \vspace{0.2cm}
    \resizebox{0.95\linewidth}{!}{
	\begin{tabular}{c|cccccc}
		\hline
		\multirow{2}{*}{\textbf{Method}} & \multirow{1}{*}{\textbf{OP}} & \multirow{1}{*}{\textbf{FLOPs}} & \multirow{1}{*}{\textbf{Params}} & \multirow{1}{*}{\textbf{Throughput}} \\ 
            & {mIoU/F1} & {(G)} & {(M)} & {(imgs/s)} \\
            \hline
            VIDNet\cite{zhou2021vid} & 0.20/0.31 & 164.7 & {82.8} & 36.9  \\ 
            FAST \cite{yu2021frequency} & 0.22/0.34 & 96.0 & 312.3 & {100.5}  \\
            HiFi-Net \cite{guo2023hierarchical} & 0.42/0.54 & 62.9 & 10.1 & 81.2  \\
            IML-ViT \cite{ma2023iml} & 0.52/0.66 & 445.6 & 88.6 & 8.8  \\
            \hline
            Mumpy-$\spadesuit$  & 0.62/0.74 & 48.0 & 262.6 & 91.2  \\ 
            Mumpy-$\clubsuit$ & 0.64/0.75 & 67.0 & 318.7 & 61.2  \\
            Mumpy-$\spadesuit\clubsuit$  & 0.59/0.71 & {29.9} & 237.9 & 98.1  \\
            Mumpy  & {0.65/0.76} &  89.2 & 361.3 & 59.1  \\ 
            \hline
	\end{tabular}
    }
    \label{tab:complexyAnalysis}
    \end{minipage}
    \begin{minipage}[c]{0.42\textwidth}
    \centering
    \hspace{-0.3cm}
    \includegraphics[width=0.33\linewidth]{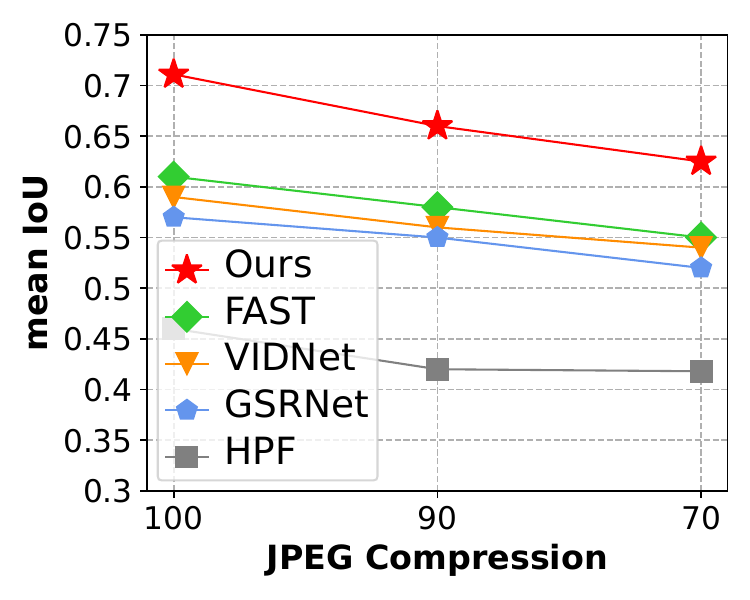}
    \hspace{-0.2cm}
    \includegraphics[width=0.33\linewidth]{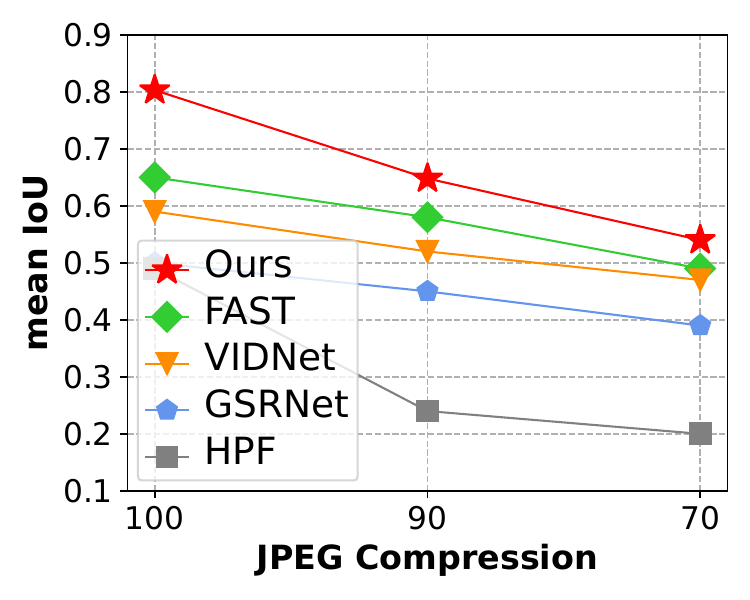}
    \hspace{-0.2cm}
    \includegraphics[width=0.33\linewidth]{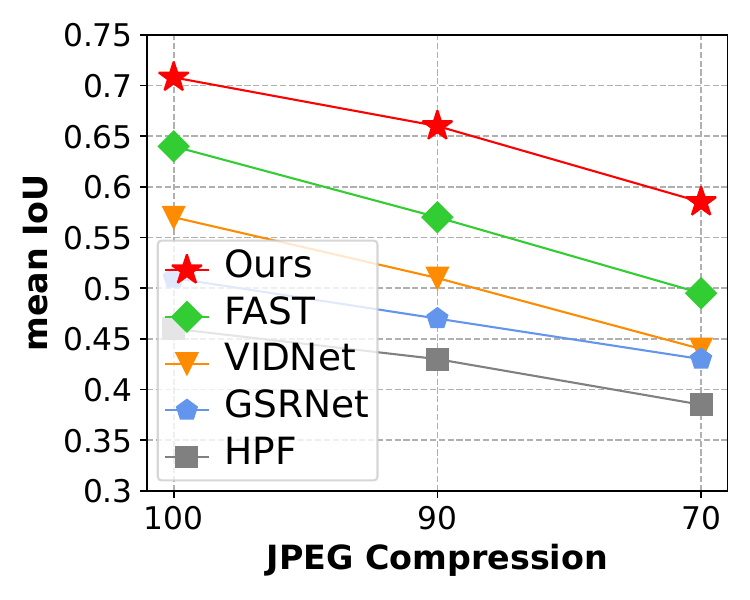}

    \includegraphics[width=0.33\linewidth]{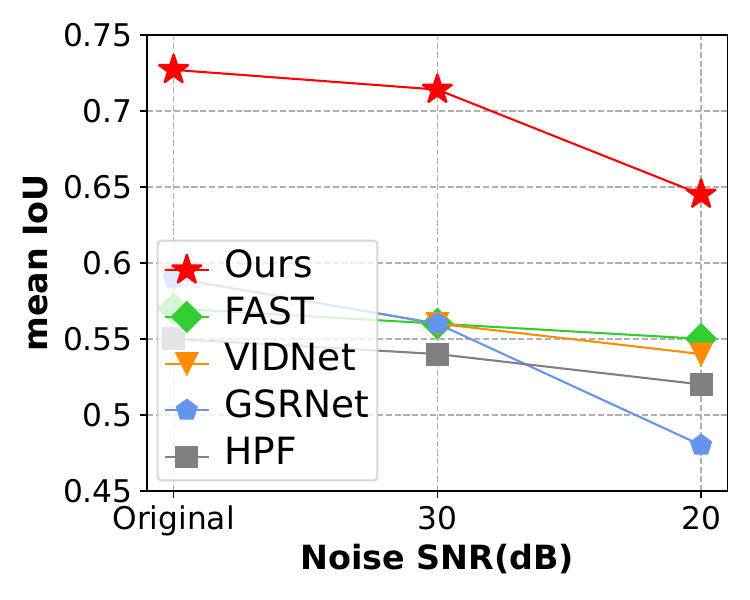}
    \hspace{-0.2cm}
    \includegraphics[width=0.33\linewidth]{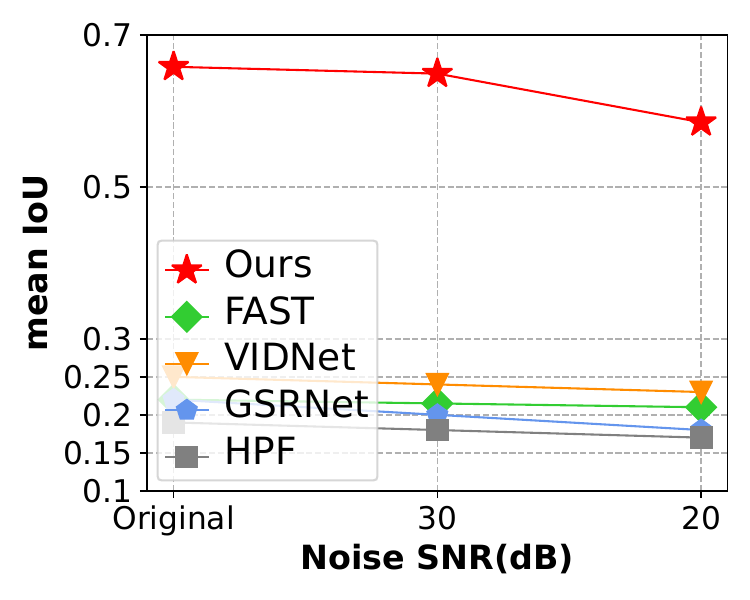}
    \hspace{-0.2cm}
    \includegraphics[width=0.33\linewidth]{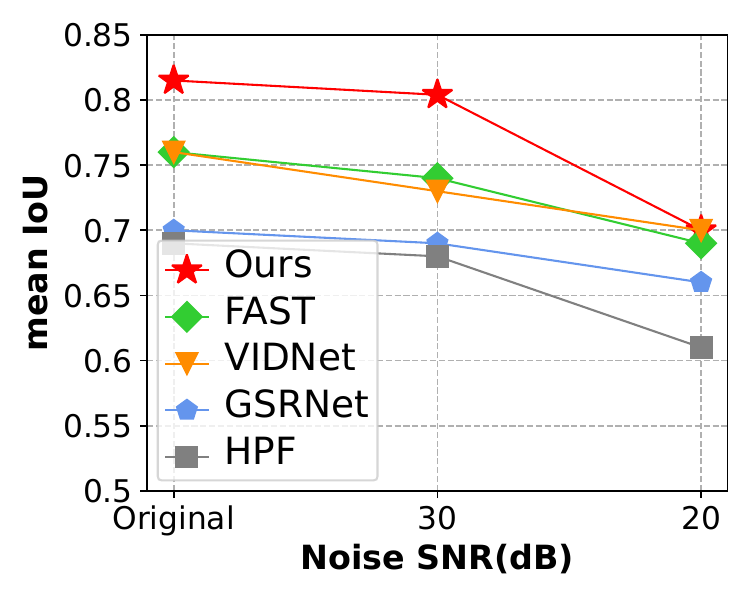}
    \vspace{0.1cm}
    \caption{\footnotesize Robustness evaluation. }
    \label{fig:robustness}
    \end{minipage}
\end{figure}

\smallskip
\noindent{\bf Ablation Study.} This study is performed on on the DVI dataset. \textbf{1) Effect of each component.} Denote \underline{Base} as only using one branch as the encoder with a traditional upsampling decoder. \underline{TF} denotes using the intermediate temporal-view features in the decoder. \underline{FF} denotes using the frequency features in the decoder. \underline{MT} denotes using the multilateral temporal-view settings. \underline{DWTI} denotes using the deformable window-based temporal-view interaction. The results in Table~\ref{tab:ablation} reveal the performance is gradually improved under both the in-domain and cross-domain scenarios by adding each component. \textbf{2) Temporal-view Selection Analysis.} We study the effect of using only one view and two views in Table~\ref{tab:viewSelection}. The results show that using all views can notably improve the cross-inpainting detection performance on OP, \ie, achieving better generalization ability.
\textbf{3) Complexity Analysis.} 
{Table~\ref{tab:complexyAnalysis} shows the computational cost and throughput of different methods. Mumpy-$\spadesuit$ denotes using Swin-Tiny for all views in encoder. Mumpy-$\clubsuit$ means using a compact decoder by downsampling all features to their 1/2. Mumpy-$\spadesuit\clubsuit$ denotes using both Swin-Tiny and compact decoder. It can be seen that despite having more parameters, our method has less FLOPs than the compared methods while showing much better cross-inpainting performance. By reducing the model size, our method still outperforms others by a large margin with much less FLOPs.}


\smallskip
\noindent{\bf Robustness Analysis.} 
Following \cite{yu2021frequency}, we train the methods on VI+OP and test them on VI, OP, and CP in the JPEG compression using the quality factors of $70$ and $90$. Moreover, we train the methods on VI+CP and test them on VI, OP, and CP in the Gaussian noise with the signal-to-noise ratios (SNR) from $20$ to $30$ dB. The results are depicted in Fig.~\ref{fig:robustness}, showing that despite all methods being degraded with increased perturbations, our method drops slower, showing the robustness of our method to resist common distortions. 

\begin{wraptable}{r}{0.4\textwidth}
\vspace{-0.4cm}
    \footnotesize
    \centering
    \caption{\small Sanity check for image-level classification AUC comparison on DVI dataset. }
    \vspace{0.2cm}
    \begin{tabular}{lccc}
    \hline
     Methods & \textbf{VI}* &\textbf{OP}* &\textbf{CP}  \\ 
     \hline
    HPF~ & 0.718 &0.640 &0.845\\
    GSRNet~ &0.762 &0.758 &0.834 \\
    VIDNet &0.778 &0.768 &0.884 \\
    FAST & 0.795 & 0.787 & 0.898\\
    OSNet & 0.992 & 0.981 & 0.989 \\
    HiFi-Net & 0.642 & 0.699 & 0.682  \\
    \rowcolor{hl}Ours & 0.996 & 0.993 & 0.997 \\
    \hline
    \end{tabular}
    \label{tab:SanityCheck}
\end{wraptable}
\smallskip
\noindent{\bf Sanity Check.}
We assess the ability of our method to distinguish between authentic and inpainted frames. We perform experiments on the DVI dataset and train methods using VI+OP, and average the pixel-wise predictions as the frame-level result. As shown in Table~\ref{tab:SanityCheck}. We can observe that our method performs best compared with others on all inpainting methods, demonstrating that our method can learn the discriminative inpainting clues. Surprisingly, HiFi-Net does not perform as expected. It may be because of the inappropriate margin setting between authentic and inpainted pixels, leading to a greater concentration on authentic pixels.

\section{Conclusion}
This paper introduces a new Transformer ({\em MumPy}) to flexibly collaborate spatial-temporal clues. To achieve this, we develop a multilateral temporal-view encoder to extract various collaborations of spatial-temporal clues and propose a deformable window-based temporal-view interaction module to enhance diversity. We then describe a multi-pyramid decoder to generate detection maps by aggregating various types of features. Our method is validated on existing datasets and a new proposed challenging Youtube-vos video inpainting dataset. The results demonstrate the efficacy of our method in both in-domain and cross-domain evaluation scenarios.

\newpage
\noindent\textbf{Acknowledgement: }
 This work was supported in part by the National Natural Science Foundation of China under Grant No.62402464, China Postdoctoral Science Foundation under Grant No.2021TQ0314 and Grant No.2021M703036.

\bibliography{egbib}

\begin{thebibliography}{41}
\providecommand{\natexlab}[1]{#1}
\providecommand{\url}[1]{\texttt{#1}}
\expandafter\ifx\csname urlstyle\endcsname\relax
  \providecommand{\doi}[1]{doi: #1}\else
  \providecommand{\doi}{doi: \begingroup \urlstyle{rm}\Url}\fi

\bibitem[Ahmed et~al.(1974)Ahmed, Natarajan, and Rao]{ahmed1974discrete}
Nasir Ahmed, T\_ Natarajan, and Kamisetty~R Rao.
\newblock Discrete cosine transform.
\newblock \emph{IEEE transactions on Computers}, 100\penalty0 (1):\penalty0 90--93, 1974.

\bibitem[Arnab et~al.(2021)Arnab, Dehghani, Heigold, Sun, Lu{\v{c}}i{\'c}, and Schmid]{arnab2021vivit}
Anurag Arnab, Mostafa Dehghani, Georg Heigold, Chen Sun, Mario Lu{\v{c}}i{\'c}, and Cordelia Schmid.
\newblock Vivit: A video vision transformer.
\newblock In \emph{Proceedings of the IEEE/CVF international conference on computer vision}, pages 6836--6846, 2021.

\bibitem[Bertasius et~al.(2021)Bertasius, Wang, and Torresani]{bertasius2021space}
Gedas Bertasius, Heng Wang, and Lorenzo Torresani.
\newblock Is space-time attention all you need for video understanding?
\newblock In \emph{ICML}, volume~2, page~4, 2021.

\bibitem[Chang et~al.(2019)Chang, Liu, Lee, and Hsu]{chang2019free}
Ya-Liang Chang, Zhe~Yu Liu, Kuan-Ying Lee, and Winston Hsu.
\newblock Free-form video inpainting with 3d gated convolution and temporal patchgan.
\newblock \emph{In Proceedings of the International Conference on Computer Vision (ICCV)}, 2019.

\bibitem[Dong et~al.(2022)Dong, Chen, Hu, Cao, and Li]{dong2022mvss}
Chengbo Dong, Xinru Chen, Ruohan Hu, Juan Cao, and Xirong Li.
\newblock Mvss-net: Multi-view multi-scale supervised networks for image manipulation detection.
\newblock \emph{IEEE Transactions on Pattern Analysis and Machine Intelligence}, 45\penalty0 (3):\penalty0 3539--3553, 2022.

\bibitem[Fan et~al.(2021)Fan, Xiong, Mangalam, Li, Yan, Malik, and Feichtenhofer]{fan2021multiscale}
Haoqi Fan, Bo~Xiong, Karttikeya Mangalam, Yanghao Li, Zhicheng Yan, Jitendra Malik, and Christoph Feichtenhofer.
\newblock Multiscale vision transformers.
\newblock In \emph{Proceedings of the IEEE/CVF international conference on computer vision}, pages 6824--6835, 2021.

\bibitem[Goodfellow et~al.(2020)Goodfellow, Pouget-Abadie, Mirza, Xu, Warde-Farley, Ozair, Courville, and Bengio]{goodfellow2020generative}
Ian Goodfellow, Jean Pouget-Abadie, Mehdi Mirza, Bing Xu, David Warde-Farley, Sherjil Ozair, Aaron Courville, and Yoshua Bengio.
\newblock Generative adversarial networks.
\newblock \emph{Communications of the ACM}, 63\penalty0 (11):\penalty0 139--144, 2020.

\bibitem[Gu et~al.(2022)Gu, Chen, Yao, Chen, Ding, and Yi]{gu2022exploiting}
Qiqi Gu, Shen Chen, Taiping Yao, Yang Chen, Shouhong Ding, and Ran Yi.
\newblock Exploiting fine-grained face forgery clues via progressive enhancement learning.
\newblock In \emph{Proceedings of the AAAI Conference on Artificial Intelligence}, volume~36, pages 735--743, 2022.

\bibitem[Guo et~al.(2023)Guo, Liu, Ren, Grosz, Masi, and Liu]{guo2023hierarchical}
Xiao Guo, Xiaohong Liu, Zhiyuan Ren, Steven Grosz, Iacopo Masi, and Xiaoming Liu.
\newblock Hierarchical fine-grained image forgery detection and localization.
\newblock In \emph{Proceedings of the IEEE/CVF Conference on Computer Vision and Pattern Recognition}, pages 3155--3165, 2023.

\bibitem[Harshvardhan et~al.(2020)Harshvardhan, Gourisaria, Pandey, and Rautaray]{harshvardhan2020comprehensive}
GM~Harshvardhan, Mahendra~Kumar Gourisaria, Manjusha Pandey, and Siddharth~Swarup Rautaray.
\newblock A comprehensive survey and analysis of generative models in machine learning.
\newblock \emph{Computer Science Review}, 38:\penalty0 100285, 2020.

\bibitem[Huang et~al.(2022)Huang, Chen, Jia, and Wang]{huang2022multi}
Shijia Huang, Yilun Chen, Jiaya Jia, and Liwei Wang.
\newblock Multi-view transformer for 3d visual grounding.
\newblock In \emph{Proceedings of the IEEE/CVF Conference on Computer Vision and Pattern Recognition}, pages 15524--15533, 2022.

\bibitem[Kim et~al.(2019)Kim, Woo, Lee, and Kweon]{kim2019deep}
Dahun Kim, Sanghyun Woo, Joon-Young Lee, and In~So Kweon.
\newblock Deep video inpainting.
\newblock In \emph{Proceedings of the IEEE/CVF Conference on Computer Vision and Pattern Recognition}, pages 5792--5801, 2019.

\bibitem[Lee et~al.(2019)Lee, Oh, Won, and Kim]{lee2019cpnet}
Sungho Lee, Seoung~Wug Oh, DaeYeun Won, and Seon~Joo Kim.
\newblock Copy-and-paste networks for deep video inpainting.
\newblock In \emph{International Conference on Computer Vision (ICCV)}, 2019.

\bibitem[Li et~al.(2021)Li, Ke, Ma, Weng, Zong, Xue, and Zhang]{li2021noise}
Ang Li, Qiuhong Ke, Xingjun Ma, Haiqin Weng, Zhiyuan Zong, Feng Xue, and Rui Zhang.
\newblock Noise doesn't lie: towards universal detection of deep inpainting.
\newblock \emph{arXiv preprint arXiv:2106.01532}, 2021.

\bibitem[Li and Huang(2019)]{li2019localization}
Haodong Li and Jiwu Huang.
\newblock Localization of deep inpainting using high-pass fully convolutional network.
\newblock In \emph{proceedings of the IEEE/CVF international conference on computer vision}, pages 8301--8310, 2019.

\bibitem[Li et~al.(2023)Li, Hu, Dong, Wu, Tian, Zhou, and Li]{li2023transformer}
Yuanman Li, Liangpei Hu, Li~Dong, Haiwei Wu, Jinyu Tian, Jiantao Zhou, and Xia Li.
\newblock Transformer-based image inpainting detection via label decoupling and constrained adversarial training.
\newblock \emph{IEEE Transactions on Circuits and Systems for Video Technology}, 2023.

\bibitem[Li et~al.(2022)Li, Lu, Qin, Guo, and Cheng]{liCvpr22vInpainting}
Zhen Li, Cheng-Ze Lu, Jianhua Qin, Chun-Le Guo, and Ming-Ming Cheng.
\newblock Towards an end-to-end framework for flow-guided video inpainting.
\newblock In \emph{IEEE Conference on Computer Vision and Pattern Recognition (CVPR)}, 2022.

\bibitem[Liu et~al.(2021{\natexlab{a}})Liu, Deng, Huang, Shi, Lu, Sun, Wang, Dai, and Li]{liu2021fuseformer}
Rui Liu, Hanming Deng, Yangyi Huang, Xiaoyu Shi, Lewei Lu, Wenxiu Sun, Xiaogang Wang, Jifeng Dai, and Hongsheng Li.
\newblock Fuseformer: Fusing fine-grained information in transformers for video inpainting.
\newblock In \emph{Proceedings of the IEEE/CVF international conference on computer vision}, pages 14040--14049, 2021{\natexlab{a}}.

\bibitem[Liu et~al.(2021{\natexlab{b}})Liu, Lin, Cao, Hu, Wei, Zhang, Lin, and Guo]{liu2021swin}
Ze~Liu, Yutong Lin, Yue Cao, Han Hu, Yixuan Wei, Zheng Zhang, Stephen Lin, and Baining Guo.
\newblock Swin transformer: Hierarchical vision transformer using shifted windows.
\newblock In \emph{Proceedings of the IEEE/CVF international conference on computer vision}, pages 10012--10022, 2021{\natexlab{b}}.

\bibitem[Ma et~al.(2023)Ma, Du, Liu, Hammadi, and Zhou]{ma2023iml}
Xiaochen Ma, Bo~Du, Xianggen Liu, Ahmed Y~Al Hammadi, and Jizhe Zhou.
\newblock Iml-vit: Image manipulation localization by vision transformer.
\newblock \emph{arXiv preprint arXiv:2307.14863}, 2023.

\bibitem[Oh et~al.(2019)Oh, Lee, Lee, and Kim]{oh2019onion}
Seoung~Wug Oh, Sungho Lee, Joon-Young Lee, and Seon~Joo Kim.
\newblock Onion-peel networks for deep video completion.
\newblock In \emph{Proceedings of the IEEE/CVF international conference on computer vision}, pages 4403--4412, 2019.

\bibitem[Park et~al.(2023)Park, Lee, and Sohn]{park2023dual}
Jungin Park, Jiyoung Lee, and Kwanghoon Sohn.
\newblock Dual-path adaptation from image to video transformers.
\newblock In \emph{Proceedings of the IEEE/CVF Conference on Computer Vision and Pattern Recognition}, pages 2203--2213, 2023.

\bibitem[Paszke et~al.(2019)Paszke, Gross, Massa, Lerer, Bradbury, Chanan, Killeen, Lin, Gimelshein, Antiga, Desmaison, Köpf, Yang, DeVito, Raison, Tejani, Chilamkurthy, Steiner, Fang, Bai, and Chintala]{paszke2019pytorch}
Adam Paszke, Sam Gross, Francisco Massa, Adam Lerer, James Bradbury, Gregory Chanan, Trevor Killeen, Zeming Lin, Natalia Gimelshein, Luca Antiga, Alban Desmaison, Andreas Köpf, Edward Yang, Zach DeVito, Martin Raison, Alykhan Tejani, Sasank Chilamkurthy, Benoit Steiner, Lu~Fang, Junjie Bai, and Soumith Chintala.
\newblock Pytorch: An imperative style, high-performance deep learning library, 2019.

\bibitem[Peng et~al.(2017)Peng, Zhang, Yu, Luo, and Sun]{peng2017large}
Chao Peng, Xiangyu Zhang, Gang Yu, Guiming Luo, and Jian Sun.
\newblock Large kernel matters--improve semantic segmentation by global convolutional network.
\newblock In \emph{Proceedings of the IEEE conference on computer vision and pattern recognition}, pages 4353--4361, 2017.

\bibitem[Perazzi et~al.(2016)Perazzi, Pont-Tuset, McWilliams, {Van Gool}, Gross, and Sorkine-Hornung]{Perazzi2016}
F.~Perazzi, J.~Pont-Tuset, B.~McWilliams, L.~{Van Gool}, M.~Gross, and A.~Sorkine-Hornung.
\newblock A benchmark dataset and evaluation methodology for video object segmentation.
\newblock In \emph{Computer Vision and Pattern Recognition}, 2016.

\bibitem[Qian et~al.(2020)Qian, Yin, Sheng, Chen, and Shao]{qian2020thinking}
Yuyang Qian, Guojun Yin, Lu~Sheng, Zixuan Chen, and Jing Shao.
\newblock Thinking in frequency: Face forgery detection by mining frequency-aware clues.
\newblock In \emph{European conference on computer vision}, pages 86--103. Springer, 2020.

\bibitem[Ren et~al.(2022)Ren, Zheng, Zhao, Xu, and Li]{ren2022dlformer}
Jingjing Ren, Qingqing Zheng, Yuanyuan Zhao, Xuemiao Xu, and Chen Li.
\newblock Dlformer: Discrete latent transformer for video inpainting.
\newblock In \emph{Proceedings of the IEEE/CVF Conference on Computer Vision and Pattern Recognition}, pages 3511--3520, 2022.

\bibitem[Sun et~al.(2023)Sun, Jiang, Wang, Li, and Cao]{sun2023safl}
Zhihao Sun, Haoran Jiang, Danding Wang, Xirong Li, and Juan Cao.
\newblock Safl-net: Semantic-agnostic feature learning network with auxiliary plugins for image manipulation detection.
\newblock In \emph{Proceedings of the IEEE/CVF International Conference on Computer Vision}, pages 22424--22433, 2023.

\bibitem[Wang et~al.(2021)Wang, Xie, Li, Fan, Song, Liang, Lu, Luo, and Shao]{wang2021pyramid}
Wenhai Wang, Enze Xie, Xiang Li, Deng-Ping Fan, Kaitao Song, Ding Liang, Tong Lu, Ping Luo, and Ling Shao.
\newblock Pyramid vision transformer: A versatile backbone for dense prediction without convolutions.
\newblock In \emph{Proceedings of the IEEE/CVF international conference on computer vision}, pages 568--578, 2021.

\bibitem[Wei et~al.(2022)Wei, Li, and Huang]{wei2022deep}
Shujin Wei, Haodong Li, and Jiwu Huang.
\newblock Deep video inpainting localization using spatial and temporal traces.
\newblock In \emph{ICASSP 2022-2022 IEEE International Conference on Acoustics, Speech and Signal Processing (ICASSP)}, pages 8957--8961. IEEE, 2022.

\bibitem[Wu et~al.(2022)Wu, Zhou, Tian, and Liu]{wu2022robust}
Haiwei Wu, Jiantao Zhou, Jinyu Tian, and Jun Liu.
\newblock Robust image forgery detection over online social network shared images.
\newblock In \emph{Proceedings of the IEEE/CVF Conference on Computer Vision and Pattern Recognition}, pages 13440--13449, 2022.

\bibitem[Xia et~al.(2022)Xia, Pan, Song, Li, and Huang]{xia2022vision}
Zhuofan Xia, Xuran Pan, Shiji Song, Li~Erran Li, and Gao Huang.
\newblock Vision transformer with deformable attention.
\newblock In \emph{Proceedings of the IEEE/CVF conference on computer vision and pattern recognition}, pages 4794--4803, 2022.

\bibitem[Xu et~al.(2018)Xu, Yang, Fan, Yang, Yue, Liang, Price, Cohen, and Huang]{xu2018youtube}
Ning Xu, Linjie Yang, Yuchen Fan, Jianchao Yang, Dingcheng Yue, Yuchen Liang, Brian Price, Scott Cohen, and Thomas Huang.
\newblock Youtube-vos: Sequence-to-sequence video object segmentation.
\newblock In \emph{Proceedings of the European conference on computer vision (ECCV)}, pages 585--601, 2018.

\bibitem[Yan et~al.(2022)Yan, Xiong, Arnab, Lu, Zhang, Sun, and Schmid]{yan2022multiview}
Shen Yan, Xuehan Xiong, Anurag Arnab, Zhichao Lu, Mi~Zhang, Chen Sun, and Cordelia Schmid.
\newblock Multiview transformers for video recognition.
\newblock In \emph{Proceedings of the IEEE/CVF conference on computer vision and pattern recognition}, pages 3333--3343, 2022.

\bibitem[Yang et~al.(2022)Yang, Zhang, Song, Hong, Xu, Zhao, Zhang, Cui, and Yang]{yang2022diffusion}
Ling Yang, Zhilong Zhang, Yang Song, Shenda Hong, Runsheng Xu, Yue Zhao, Wentao Zhang, Bin Cui, and Ming-Hsuan Yang.
\newblock Diffusion models: A comprehensive survey of methods and applications.
\newblock \emph{ACM Computing Surveys}, 2022.

\bibitem[Yu et~al.(2021)Yu, Li, Li, Lu, and Zhou]{yu2021frequency}
Bingyao Yu, Wanhua Li, Xiu Li, Jiwen Lu, and Jie Zhou.
\newblock Frequency-aware spatiotemporal transformers for video inpainting detection.
\newblock In \emph{Proceedings of the IEEE/CVF International Conference on Computer Vision}, pages 8188--8197, 2021.

\bibitem[Zhang et~al.(2020)Zhang, Zhang, Tang, Wang, Hua, and Sun]{zhang2020feature}
Dong Zhang, Hanwang Zhang, Jinhui Tang, Meng Wang, Xiansheng Hua, and Qianru Sun.
\newblock Feature pyramid transformer.
\newblock In \emph{Computer Vision--ECCV 2020: 16th European Conference, Glasgow, UK, August 23--28, 2020, Proceedings, Part XXVIII 16}, pages 323--339. Springer, 2020.

\bibitem[Zhang et~al.(2022)Zhang, Fu, and Liu]{Zhang_2022_CVPR}
Kaidong Zhang, Jingjing Fu, and Dong Liu.
\newblock Inertia-guided flow completion and style fusion for video inpainting.
\newblock In \emph{Proceedings of the IEEE/CVF Conference on Computer Vision and Pattern Recognition (CVPR)}, pages 5982--5991, June 2022.

\bibitem[Zhou et~al.(2020)Zhou, Chen, Han, Najibi, Shrivastava, Lim, and Davis]{zhou2020generate}
Peng Zhou, Bor-Chun Chen, Xintong Han, Mahyar Najibi, Abhinav Shrivastava, Ser-Nam Lim, and Larry Davis.
\newblock Generate, segment, and refine: Towards generic manipulation segmentation.
\newblock In \emph{Proceedings of the AAAI conference on artificial intelligence}, volume~34, pages 13058--13065, 2020.

\bibitem[Zhou et~al.(2021)Zhou, Yu, Wu, Davis, Shrivastava, and Lim]{zhou2021vid}
Peng Zhou, Ning Yu, Zuxuan Wu, Larry~S Davis, Abhinav Shrivastava, and Ser~Nam Lim.
\newblock Deep video inpainting detection.
\newblock In \emph{BMVC}, 2021.

\bibitem[Zhou et~al.(2023)Zhou, Li, Chan, and Loy]{zhou2023propainter}
Shangchen Zhou, Chongyi Li, Kelvin~C.K Chan, and Chen~Change Loy.
\newblock {ProPainter}: Improving propagation and transformer for video inpainting.
\newblock In \emph{Proceedings of IEEE International Conference on Computer Vision (ICCV)}, 2023.

\end{thebibliography}
\end{document}